\RequirePackage{fix-cm}
\documentclass{amsart}

\usepackage[pdftitle={SUPRA: Open Source Software Defined Ultrasound Processing for Real-Time Applications}, pdfauthor={Ruediger Goebl, Nassir Navab, Christoph Hennersperger }, colorlinks=true]{hyperref}
\usepackage[all]{hypcap}        % needed to help hyperlinks direct correctly;
\usepackage{subfig}
\usepackage{graphicx}
\usepackage{color}
\usepackage{siunitx}
\usepackage{rotating}
\usepackage{multirow}
\usepackage[foot]{amsaddr}

\newcommand{\step}[1]{(#1)}
\newcommand{\shortcaption}[1]
  {\textbf{#1}}
	
\newcommand{\mysubref}[1]
  {\textbf{(#1)}}
\newcommand{\meanStd}[2]{$#1 \pm \scriptstyle#2$ }
\begin{document}

\title[SUPRA: Software Defined Ultrasound Processing]{SUPRA: Open Source Software Defined Ultrasound Processing for Real-Time Applications\\\small{A 2D and 3D Pipeline from Beamforming to B-mode}}
%Grants or other notes
%about the article that should go on the front page should be
%placed here. General acknowledgments should be placed at the end of the article.}

\thanks{This project has received funding from the European Union's Horizon 2020 research and innovation program EDEN2020 under grant agreement No 688279}
%\subtitle{A 2D and 3D Pipeline from Beamforming to B-mode}

%\titlerunning{SUPRA: Software Defined Ultrasound Processing for Real-Time Applications}        % if too long for running head

\author{R\"udiger G\"obl}
\address[R. G\"obl, C. Hennersperger]{Computer Aided Medical Procedures, Technische Universit\"at M\"unchen, Boltzmannstr. 3, 85748 Garching, Germany}
\email{r.goebl@tum.de}
\thanks{}

%    author two information
\author{Nassir Navab}
\address[Nassir Navab]{Computer Aided Medical Procedures, Technische Universit\"at M\"unchen, Boltzmannstr. 3, 85748 Garching, Germany and 
			Johns Hopkins University, 3400 North Charles Street, Baltimore, MD 21218, USA}

\author{Christoph Hennersperger}


\date{2017-11-16}

\dedicatory{}

%\authorrunning{Short form of author list} % if too long for running head

%\institute{R\"udiger G\"obl \and
%    		Christoph Hennersperger \at
%            Computer Aided Medical Procedures, Technische Universit\"at M\"unchen, Boltzmannstr. 3, 85748 Garching, Germany \\
%            Tel.: +49 89 289 19412\\
%            \email{r.goebl@tum.de}           %  \\
%             \emph{Present address:} of F. Author  %  if needed
%			\and
%			Nassir Navab \at
%			Computer Aided Medical Procedures, Technische Universit\"at M\"unchen, Boltzmannstr. 3, 85748 Garching, Germany and 
%			Johns Hopkins University, 3400 North Charles Street, Baltimore, MD 21218, USA
%}

%\date{Received: date / Accepted: date}
% The correct dates will be entered by the editor

\begin{abstract}
Research in ultrasound imaging is limited in reproducibility by two factors: First, many existing ultrasound pipelines are protected by intellectual property, rendering exchange of code difficult. Second, most pipelines are implemented in special hardware, resulting in limited flexibility of implemented processing steps on such platforms.
\newline
\textbf{Methods}
With SUPRA we propose an open-source pipeline for fully Software Defined Ultrasound Processing for Real-time Applications to alleviate these problems. Covering all steps from beamforming to output of B-mode images, SUPRA can help improve the reproducibility of results and make modifications to the image acquisition mode accessible to the research community. We evaluate the pipeline qualitatively, quantitatively, and regarding its run-time.
\newline
\textbf{Results}
The pipeline shows image quality comparable to a clinical system and backed by point-spread function measurements a comparable resolution. Including all processing stages of a usual ultrasound pipeline, the run-time analysis shows that it can be executed in 2D and 3D on consumer GPUs in real-time.
\newline
\textbf{Conclusions}
Our software ultrasound pipeline opens up the research in image acquisition. Given access to ultrasound data from early stages (raw channel data, radiofrequency data) it simplifies the development in imaging. Furthermore, it tackles the reproducibility of research results, as code can be shared easily and even be executed without dedicated ultrasound hardware.
% \PACS{PACS code1 \and PACS code2 \and more}
% \subclass{MSC code1 \and MSC code2 \and more}
\end{abstract}

\maketitle

\section{Introduction}
\label{sec:intro}

Ultrasound (US) imaging is used in a wide variety of applications and complements modalities such as computed tomography (CT) and magnetic resonance imaging (MRI) by the information it provides and through the ways it can be used. 
As with MRI, Ultrasound can be used to gather anatomical, dynamical, as well as, functional information.
This property combined with its uncumbersome use allow for US imaging to be used during interventions with little modifications to the surgeon's workflow.
One of the main advantages of US is its non-invasive nature, as only compressional waves with low intensity and power are used.
In addition to that, the costs of ultrasound systems are low compared to CT and MRI. Finally, US imaging devices can be fully portable, allowing seamless use directly at the bedside.
Because of this, there is a trend to replace MRI and especially CT with US imaging when applicable.

The image acquisition process of US can be broken down into several steps forming a pipeline.
\begin{itemize}
\item First, an ultrasonic pulse modulated with a chosen frequency is generated in the tissue through precisely orchestrated electrical excitation of piezo transducer elements \step{transmit beamforming}.
\item The echos induced by this pulse in the tissue are converted to electrical signals by the transducer elements and commonly stored digitally \step{receive}.
\item After that, the signals are used to compute how they would have been received from one single line, where the data from different channels is delayed such that scattered signals from that line are intensified by constructive interference and echoes from other positions are reduced through destructive interference \step{receive beamforming} \cite{Shattuck1984}.
\item The result of this operation is called the radio frequency (RF) data, because the echos are still modulated with the transmit frequency.
\item Since only the scattered intensity is of interest to the user, the RF-data is demodulated \step{envelope detection}, leaving the signal amplitude.
\item Before these amplitudes are shown on a screen they undergo a non-linear compression stage to match their dynamic range to the perceived dynamic range \step{log-compression}.
\item Finally, the single lines that have been reconstructed and which are not necessarily parallel to each other are interpolated to create an image representing physical dimensions \step{scan conversion}.
\end{itemize}
In current US systems, those steps are usually performed in specialized hardware, field programmable gate arrays (FPGAs), or specialized processors such as digital signal processors (DSPs).
Implementations like this have been a necessity in the early years of ultrasound. With the increase of computing power available in modern workstations, there is no longer a strong need for that, especially when using GPUs to perform the numerical calculations.
When implemented in separate hardware, modifications to the algorithms are hardly possible, as programming interfaces might not be available and the development for FPGAs and DSPs is highly complex and time-consuming.
This makes the development of new methods cumbersome and essentially impedes research on ultrasound imaging.
Especially for research applications, such limitations often lead to the utilization of frame-grabbers to allow for a retrieval of US images from clinical scanners \cite{Zettinig2015} even in recent publications \cite{riva20173d,bluemel2016fusion}, which is not only limiting reproducibility but also potentially hampers image quality. Problematic influences on the image quality include lossy image compression, frame rate differences between the frame-grabber and the US acquisition and lines, text and markers superimposed on the images.

There already exist some ultrasound systems, in which all of the processing already happens on GPUs, thus allowing the manufacturer of the device to implement changes to their pipeline more rapidly.
Yet, as vendors protect their intellectual property, it is difficult for independent research groups to use those machines for research purposes.
On the other hand, there have been efforts to give researchers access to parts of the US pipelines they employ, as for example by the PLUS toolkit\cite{Lasso2014}.
While this project is tailored to tracked and freehand 3D-ultrasound applications, it does not provide US beamforming or low-level processing capabilities. 
Besides the efforts around the PLUS toolkit, a group of researchers recently announced the UltraSound ToolBox for MATLAB \cite{Rodriguez-Molares2017}. While this platform provides basic capabilities for beamforming and development of advanced low-level processing methodologies, it is conceptionally not directed towards the use in real-time applications, thus not being suited for many applications in computer assisted interventions.

On this basis, we try to close the gap between low-level \emph{offline} ultrasound research on the one side, and the processing of \emph{already processed} US images on the other. We propose an open source pipeline for 2D and 3D ultrasound imaging called SUPRA (Software defined Ultrasound Processing for Real-time Applications) that can be used to perform all computation-based steps in US and enable researchers and developers to work on all parts of the imaging process. This ranges from beamforming the raw channel data recorded by an US-system to the output of B-mode images.
Thus, we hope to integrate efforts of other platforms, and also specifically provide a way towards a stronger integration of high-level processing (targeting e.g. at a specific medical image computing or computer aided intervention application) with low-level, ultrasound-specific information (e.g. raw channel or RF data). 
This way, SUPRA could for example be used to integrate efforts from image segmentation throughout all levels of the US processing pipeline, potentially leveraging specific first order data not considered so far.
\section{SUPRA}
\label{sec:supra}

In the following, we describe our approach towards Software Defined Ultrasound Processing for Real-time Applications (SUPRA). The framework is licensed under LGPLv3 and designed as cross-platform solution tested with both Windows and Linux/Ubuntu. It is publicly available on GitHub\footnote{\url{https://github.com/IFL-CAMP/supra}}.

The term software defined ultrasound is derived from a concept called ``software defined radio''. In this field of radio communication, hardware implementations of signal processing components such as filters, amplifiers and modulators are replaced by software implementations. This can help to reduce costs and simplify their development.

\begin{figure}[htb]
	\centering
	\includegraphics[width=0.8\textwidth]{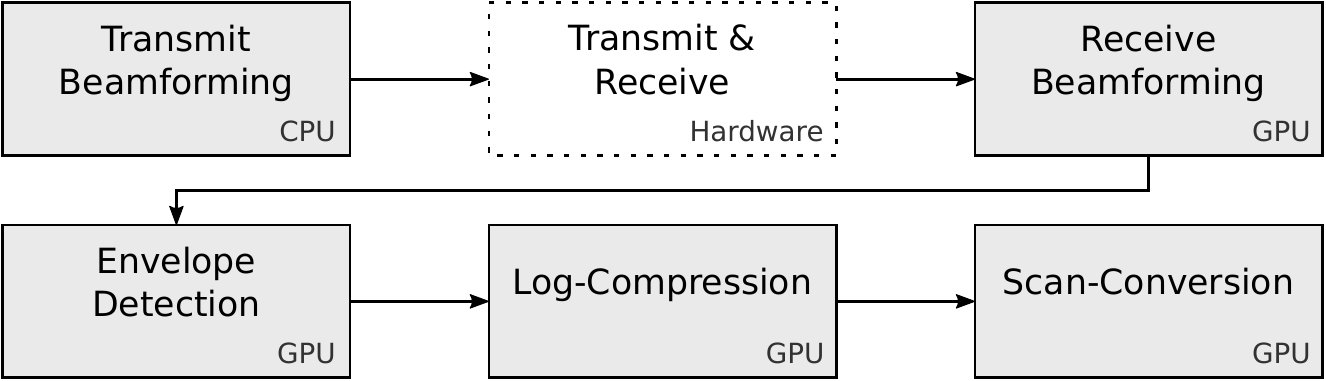}
	\caption{\shortcaption{Implemented pipeline} with the modules realized in SUPRA, hence in software, marked with solid lines and the module that has to be performed in hardware with dashed lines. 
	\label{fig:pipeline-overview}
	}
\end{figure}

Our framework SUPRA follows the same concept.
In addition to being fully implemented in software, all the real-time critical processing steps have been implemented in NVIDIA CUDA to achieve high throughput. 
\autoref{fig:pipeline-overview} shows a pipeline with the steps as outlined in Section~\ref{sec:intro} and highlights where the respective processing takes place.
The transmit beamforming is performed on the CPU, as it is only necessary to compute the transmit parameters once for a fixed acquisition. 
It is important to note here that \emph{only} the actual transmit and recieve steps require specific hardware with an analogue frontend, allowing for the excitation of piezo-elements in the US probe to create acoustic waves.
Thus, all other steps in the pipeline (receive beamforming, envelope detection, log-compression, and scan conversion) can be fully customized in software.
Since these are performed repeatedly, we implemented them in NVIDIA CUDA.
As a consequence, it is possible to execute the pipeline for 3D ultrasound on consumer graphics cards, as we show in Subsection~\ref{subsec:eval-performance}.

Besides the aim to maximize real-time capability by parallelizing relevant parts of the ultrasound pipeline, we also considered the modularity of the pipeline as a major design goal.
We achieved this by encapsulating the processing components into nodes on a data-flow graph, realized with the Intel Thread Building Blocks open source library\footnote{\url{https://www.threadingbuildingblocks.org/}}.
Based on this architecture, the nodes only exchange shared pointers to data containers (which may reside either on the CPU or the GPU) among each other, eliminating unnecessary memory operations.

In detail, each encapsulated node can utilize an input, provide an output, or both. On this foundation, by placing the processing steps of the US pipeline within the data-flow graph, nodes can be added, removed, or exchanged without recompiling the actual code. 
Besides this possibility to exchange individual parts of the pipeline, the overall pipeline can also be altered completely using the input-output mechanisms. In this way, it is for example possible to perform two differently parametrized beamforming runs in parallel on the same input data to extract different information. In a hardware-based pipeline this is not possible, while with SUPRA such considerations are limited only by the computational power.

In view of our efforts to provide a fully functional basic ultrasound beamforming and processing pipeline, the following methods are available in SUPRA:
\begin{enumerate}
    \item Dynamic transmit and receive beamforming for a fully flexible scanline layout and resolution, as well as full control of acoustic wave excitations and multi-line acquisitions
    \item Delay and sum beamforming for received raw channel data directly after analogue to digital conversion \cite{Shattuck1984}
    \item Envelope detection by IQ-demodulation, including frequency compounding through a bank of configurable bandpasses
    \item Configurable log compression for target imaging dynamic range
    \item Scan conversion in 2D and 3D, for a wide range of scan-line configurations
    \item Graphical user interface for online-configuration and real-time visualization of received data
    \item Configurable XML-interface for the generation of system parameters and specific imaging pipelines
\end{enumerate}

In addition to the nodes that make up the core pipeline, several input and output nodes are present.
Input nodes provide the interface to the actual ultrasound system hardware (i.e. hardware transmit and receive as indicated above) and are thus vendor or system-specific interface implementations. 
At this stage, Cephasonics ultrasound hardware (Cephasonics, Santa Clara, CA, USA) is integrated with the full pipeline for beamforming; and Ultrasonix systems (BK Ultrasound, Peabody, MA, USA) can be interfaced using the proprietary ulterius software interface. It should be noted, however, that the integration of other hardware-platforms would only require the implementation of a new input node within the data graph, providing a respective interface to the system-specific transmit and receive hardware.

Output nodes provide a way to either forward information at any stage of the overall pipeline to a dedicated interface, or provide a means to save data to a file on the hard-disk respectively. Implemented output nodes represent at this stage:
\begin{itemize}
    \item ROS bridge for interaction with robotic environments,
    \item OpenIGTLink bridge for exchange of image and tracking data,
    \item Storage of information as meta images (mhd) for offline use.
\end{itemize}
It is worth noting that output nodes are not limited to the last step of the pipeline, but can also be used with any intermediate data stream present in the system.
Following the generic and modular software architecure, nodes for input and output can also contain other information. For the use in interventional settings, tracking information can be attached to the images.
In this view, SUPRA provides generic interfaces for tracking in- and output via OpenIGTLink and as ROS messages. Additionally, there are native interfaces to Ascencion EM trackers, as well as NDI optical trackers.

\begin{figure}[htb]
	\centering
	\includegraphics[width=0.95\textwidth]{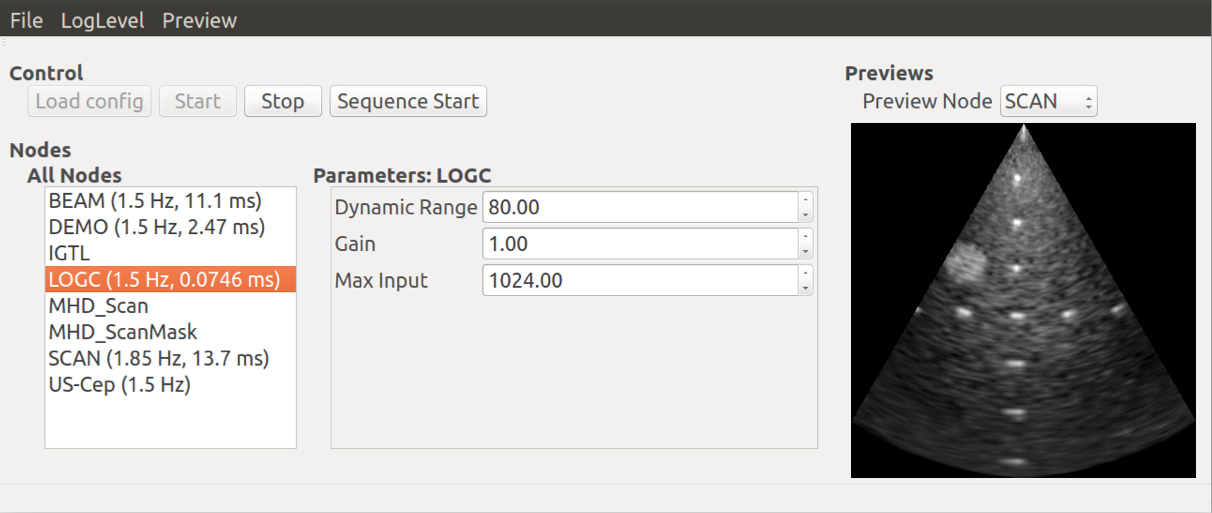}
	\caption{\shortcaption{Graphical user interface} with the processing nodes in the left column, the parameters of the selected node in the central column and a live preview of one data stream on the right.
	\label{fig:gui}
	}
\end{figure}
The online-configuration mentioned before is realized via a generic parameter system, that enables all processing nodes to define parameters with default values and valid value ranges (continuous, discrete), where applicable.
Through this parameter-system the respective nodes are notified of parameter changes and can react accordingly.
The parameters can be inspected and modified during run-time by the user in the GUI, which is shown in \autoref{fig:gui}.
Additionally, for the use in automated systems, the parameters can be accessed through a ROS service, allowing fully dynamic imaging.

To this end, in order to use the full pipeline as introduced above, a respective US system is required, providing access to the beamforming parameters and the raw data collected at channel level has to be available. However, even without this access, a user can still apply the other parts of the pipeline to harness the full control over those steps, e.g. by injecting previously acquired data into the pipeline.

Additionally, with the planned inclusion of image post-processing to SUPRA, such as speckle reduction techniques \cite{Krissian2007}, or advanced imaging protocols like harmonic imaging \cite{averkiou2000} and planewave imaging \cite{tanter2014ultrafast}, the current baseline provided by the framework will allow researchers to evaluate their methods in a more meaningful manner.
\section{Comparison}
\label{sec:comparison}
For the evaluation in this work, we employ the proposed SUPRA pipeline with a 384 channel cQuest Cicada™ from Cephasonics, CA, USA and a Cephasonics CPLA12875 transducer with \SI{7}{MHz} center frequency, 128 elements, and \SI{0.3}{mm} pitch and work with the raw channel data collected by this system. For this transducer only 64 channels can be used.
The resulting images are compared qualitatively as well as quantitatively to a clinical Ultrasonix® Sonix RP US system (BK Ultrasound, Peabody, MA, USA) equipped with a L14-5/38 linear transducer (128 elements, \SI{0.3}{mm} pitch).
Both systems use the following parameters: frequency: \SI{6.6}{MHz}, depth: \SI{4.5}{mm}.
Additionally, we show acquisitions performed with a 2D matrix probe with all 384 channels of the Cephasonics interface.

\subsection{Qualitative comparison}
\label{subsec:qualitative}
For a qualitative comparison of the image quality between the two systems, we show in-vivo images of the carotid artery and the biceps tendon in combination with the brachialis of a healthy volunteer, as well as phantom data acquired with a CIRS multi purpose phantom (Model 040GSE). \autoref{fig:images-qualitative} shows in the first and second row cross-sectional and longitudinal views of a common carotid artery, where images acquired with the Sonix RP are on the left and those created with SUPRA on the right.
Due to tissue deformations and limited probe placement reproducibility, the anatomy shown is not exactly the same. Despite this, it can be observed, that while for the cross-section SUPRA shows less clutter noise inside the carotid than the Sonix RP, this is reversed in the longitudinal view.
Apart from that, both systems show comparable tissue texture in the muscle covering the carotid, although it appears slightly less blurred with the Sonix RP.
In the tendon and muscle depicted in the third row, the previous observation can also be made.
The mentioned blur is likely caused by settings of the frequency compounding during envelope detection, as the filters used to separate different frequencies of the RF lines can cause blurring.
Despite this potential shortcoming of SUPRA, it is necessary to point out, that the clinical system employs further processing steps after scan-conversion, such as speckle reduction, which are not implemented in SUPRA presently.
In the fourth row of \autoref{fig:images-qualitative} there are images of a CIRS multi purpose phantom (Model 040GSE).
In contrast to the in-vivo images, here the tissue texture in the image acquired with SUPRA appears less blurred than with the Sonix RP.
Additionally, the wires in the lower part of the images exhibit a higher contrast in SUPRA and seem less blurred as well.
The qualitative comparison thus shows, that the image quality of SUPRA is comparable to what can be achieved with the Sonix RP.

\def\thisfigwidth{0.336\linewidth}
\begin{figure*}
	\centering
	\captionsetup[subfigure]{labelformat=empty}
	\setlength{\tabcolsep}{2pt}
	\renewcommand{\arraystretch}{0}
	\begin{tabular}{lcc}
		\raisebox{.145\linewidth}[0pt][0pt]{\rotatebox[origin=c]{90}{\centering Carotid cross-sect.}}&
		\subfloat[]{
		    \includegraphics[width=\thisfigwidth,clip=true,trim=0 200 0 0]{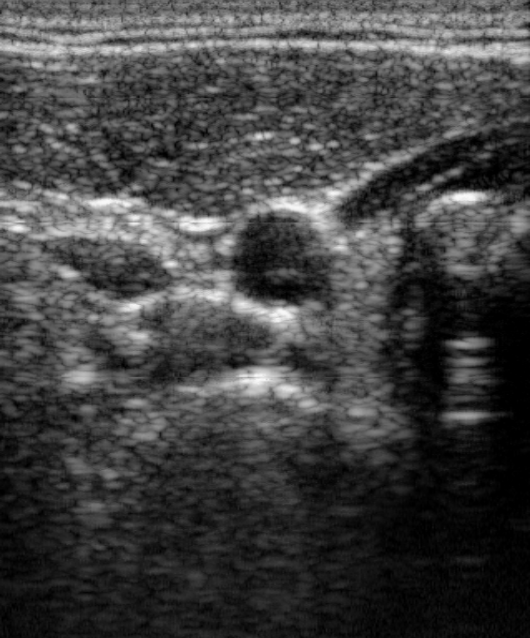}
		    \label{fig:image-carotid-cross-ultrasonix}}&
		\subfloat[]{
		    \includegraphics[width=\thisfigwidth,clip=true,trim=0 365 0 10]{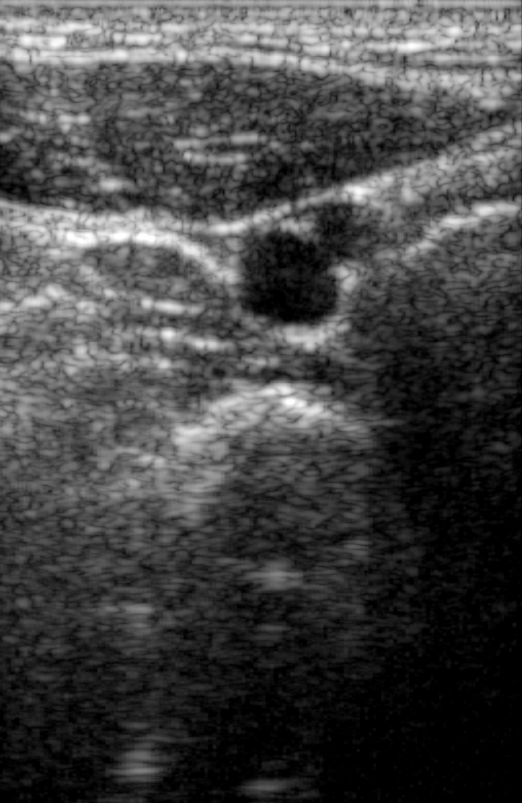}
		    \label{fig:image-carotid-cross-supra}}
		\\[-2em]
		\raisebox{.155\linewidth}[0pt][0pt]{\rotatebox[origin=c]{90}{\centering Carotid long.}}&
		\subfloat[]{
		    \includegraphics[width=\thisfigwidth,clip=true,trim=0 190 0 0]{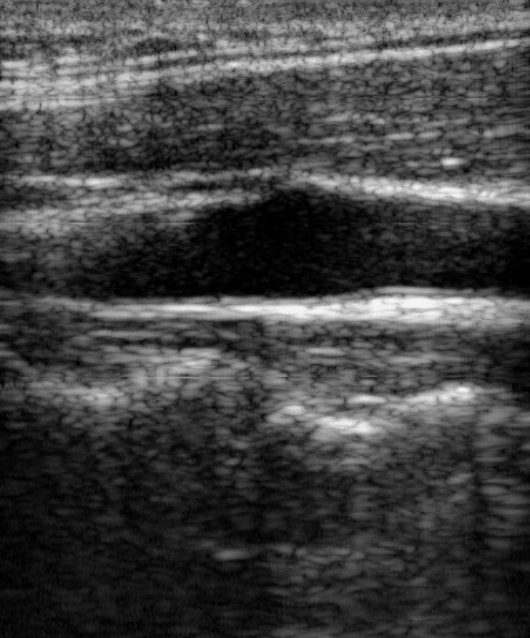}
		    \label{fig:image-carotid-long-ultrasonix}}&
		\subfloat[]{
		    \includegraphics[width=\thisfigwidth,clip=true,trim=0 355 0 10]{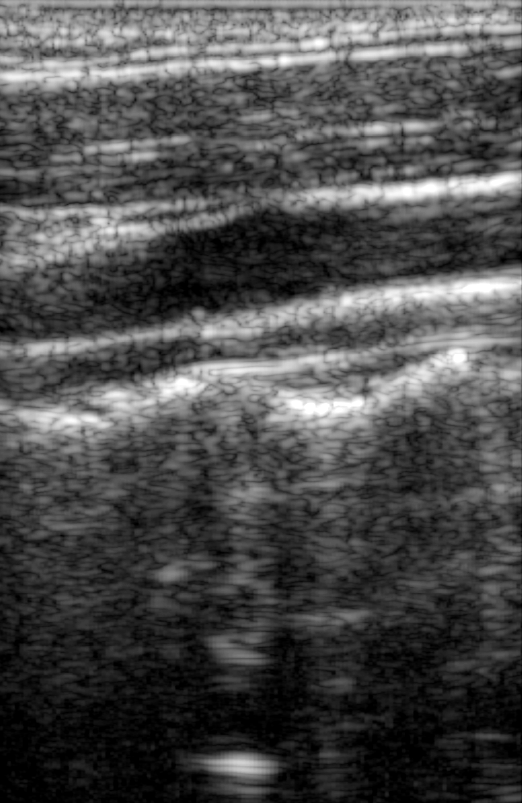}
		    \label{fig:image-carotid-long-supra}}
		\\[-2em]
		\raisebox{.18\linewidth}[0pt][0pt]{\rotatebox[origin=c]{90}{\centering Biceps tendon and bracialis long.}}&
		\subfloat[]{
		    \includegraphics[width=\thisfigwidth,clip=true,trim=0 70 0 0]{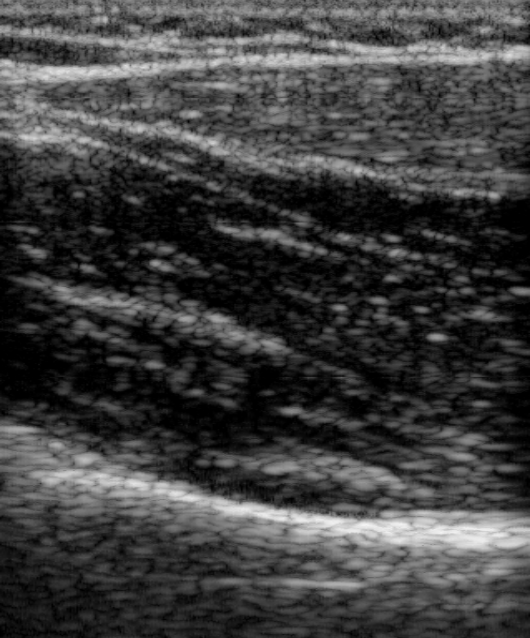}
		    \label{fig:image-biceps-long-ultrasonix}}&
		\subfloat[]{
		    \includegraphics[width=\thisfigwidth,clip=true,trim=0 235 0 10]{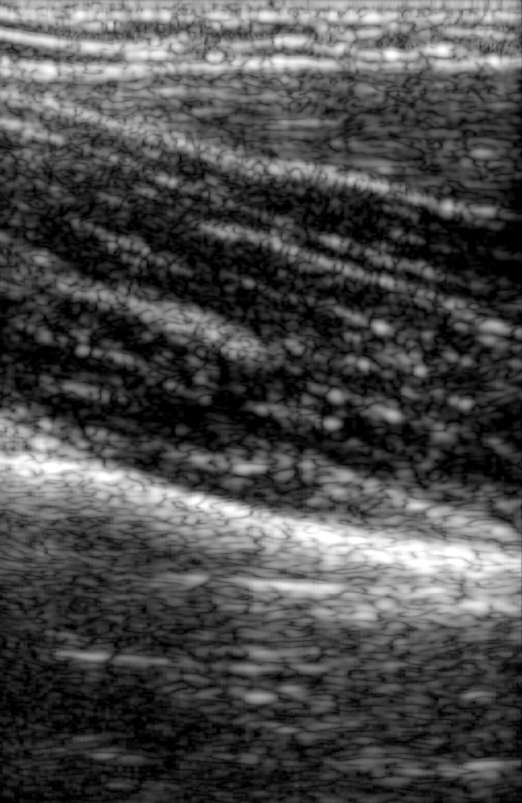}
		    \label{fig:image-biceps-long-supra}}
		\\[-2em]
		\raisebox{.175\linewidth}[0pt][0pt]{\rotatebox[origin=c]{90}{\centering CIRS phantom}} &
		\subfloat[]{
		    \includegraphics[width=\thisfigwidth,height=\thisfigwidth,clip=true,trim=0 20 0 100]{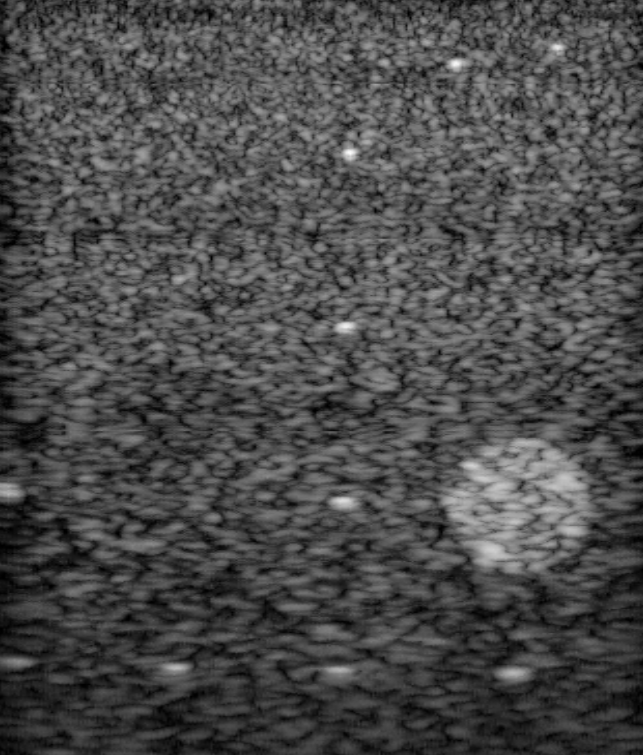}
		    \label{fig:image-cirs-ultrasonix}}&
		\subfloat[]{
		    \includegraphics[width=\thisfigwidth,height=\thisfigwidth,clip=true,trim=0 35 0 125]{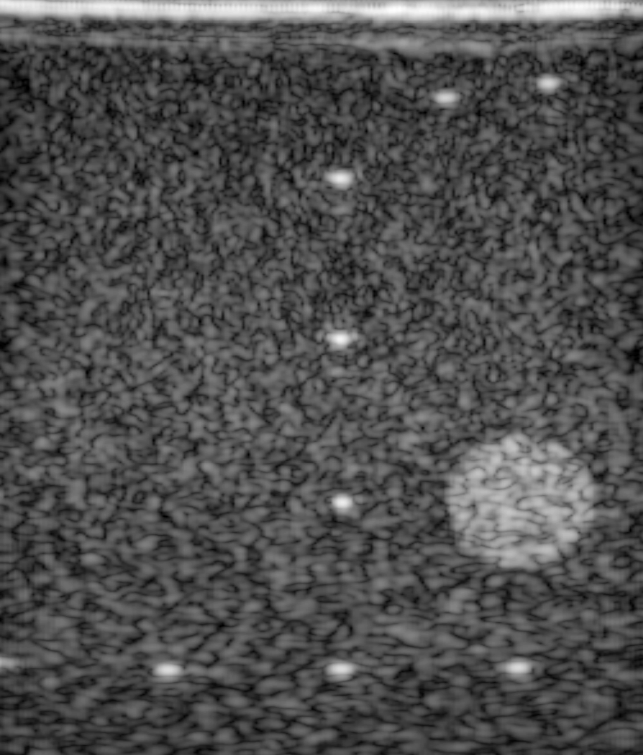}
		    \label{fig:image-cirs-supra}}
		\\
		\\[-0.9em]
		& 
		Ultrasonix &
		SUPRA
	\end{tabular}
	\caption{\shortcaption{Qualitative comparison} of Ultrasonix Sonix RP US (left) and SUPRA with a Cephasonics cQuest Cicada equipped with a linear transducer at \SI{6.6}{MHz}, depth \SI{45}{mm}. The first three rows show in-vivo acquisitions of the carotid (cross-sectional and longitudinal) and the biceps / brachialis of a healthy volunteer. The last row shows images of a CIRS multi purpose phantom (Model 040GSE).}
	\label{fig:images-qualitative}
\end{figure*}

\begin{figure}[htb]
	\centering
	%\begin{tabular}{c|c}
	\subfloat[]{
		\includegraphics[width=\thisfigwidth,clip=true,trim=0 0 0 0]{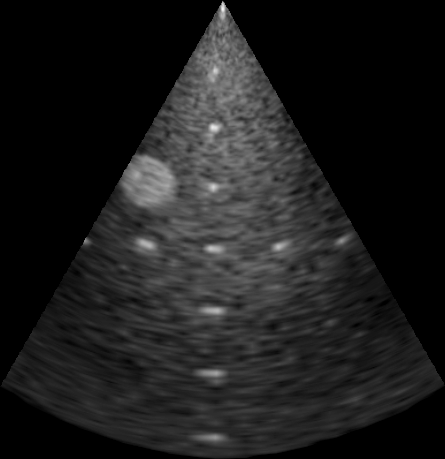}
		\label{fig:3d_slices_a}
		}
	\quad
	\subfloat[]{
		\includegraphics[width=\thisfigwidth,clip=true,trim=0 0 0 0]{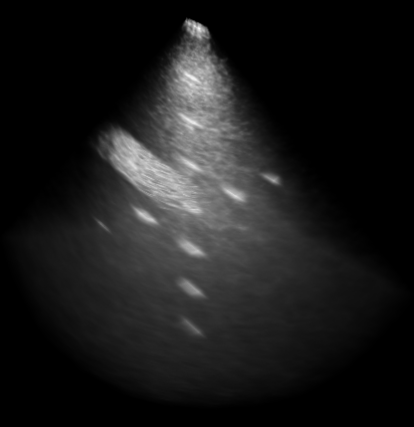}
		\label{fig:3d_slices_b}
		}
	\\
	\subfloat[]{
		\includegraphics[width=\thisfigwidth,clip=true,trim=0 0 0 0]{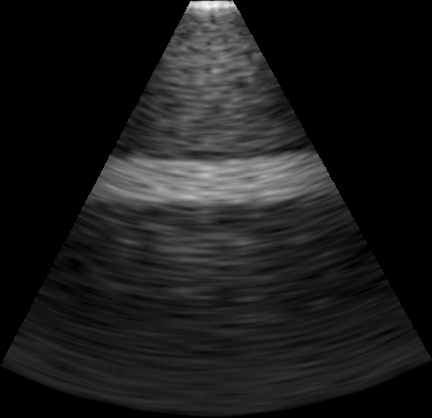}
		\label{fig:3d_slices_c}
		}
	\quad
	\subfloat[]{
		\includegraphics[width=\thisfigwidth,clip=true,trim=0 0 0 0]{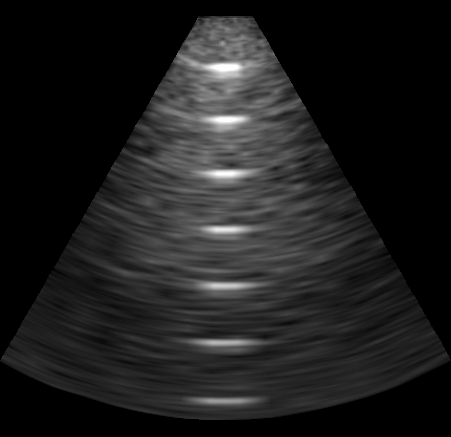}
		\label{fig:3d_slices_d}
		}
	\caption{\shortcaption{3D US volume acquired with SUPRA} of a CIRS multi purpose phantom (Model 040GSE). \mysubref{a} shows a cross-section of the structures in the phantom. A rendering of the volume is shown in \mysubref{b}. The second row shows two longitudinal views, \mysubref{c} of a hyperechoic region, \mysubref{d} wires. Note the limited visibility of the horizontal wires in \mysubref{d}, caused by their highly specular reflectivity.
	\label{fig:3d_slices}
	}
	%\end{tabular}
\end{figure}

As pointed out earlier, SUPRA is not limited to classical 2D ultrasound. It is also capable of performing all processing steps for data acquired with 3D ultrasound probes, also known as 2D array or matrix probes.
Slices of a volume acquired with a Vermon matrix probe are shown in \autoref{fig:3d_slices}. The probe contains an array of $32 \times 32$ elements and was driven by SUPRA via a Cephasonics cQuest Cicada with 384 channels. The volume shown was acquired with \SI{70}{mm} depth, a frequency of \SI{7}{MHz} and with 512 ($32 \times 16$) scanlines over a field of view of \SI{60}{\degree}. It depicts parts of the CIRS multi purpose phantom. The top-left image shows a volume slice perpendicular to all internal structures and it is clearly visible that the resolution decreases with increasing depth, as is expected for a phased array. The image in the top-right shows a 3D rendering of the volume, where the different lengths of the the hyperechoic inclusion and the wires are apparent.
Bottom left shows a slice perpendicular to that, including a longitudinal view of the hyperechoic inclusion and the image in the bottom-right shows a longitudinal view of the horizontal wires. It is worth noting that the wires are longer than is visible from the slice, but due to the increasing scanline angles to the image boundaries natural to phased arrays in combination with the highly specular reflectivity of the wires, their visibility quickly falls off with distance from the center.

\subsection{Quantitative evaluation}
To complement the qualitative comparison with a quantitative evaluation of both beamformers, we estimate the point-spread functions (PSFs) of both systems, following the approach of Jeong \cite{Jeong2015}.
For this purpose, we imaged a wire target in a tank with distilled water at \SI{47}{\degree C} at different depths, while acquiring the RF data after beamforming, both for the Ultrasonix Sonix RP and SUPRA.
Afterwards we performed envelope detection through the hilbert transform in a numerics software followed by a log compression to a dynamic range of \SI{50}{dB}.

\begin{figure}[htb]
	\centering
	%\begin{tabular}{c|c}
	%\subfloat*[]{
		\includegraphics[width=0.9\textwidth,clip=true,trim=35 25 42 35]{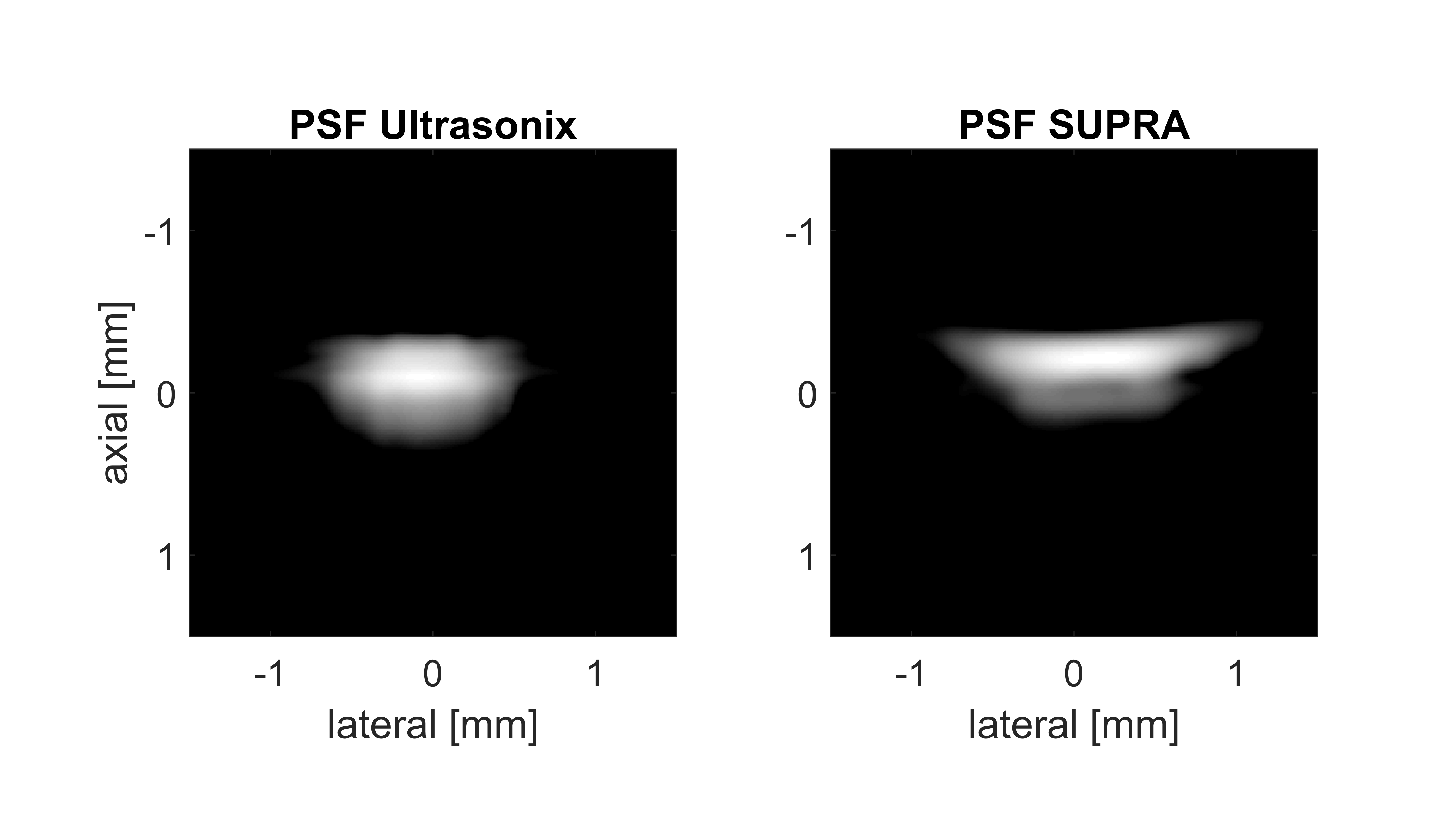}
	%	\label{fig:psf-comparison-a}
	%	}
	\\[1em]
	%\subfloat*[]{
	    \includegraphics[width=0.9\textwidth]{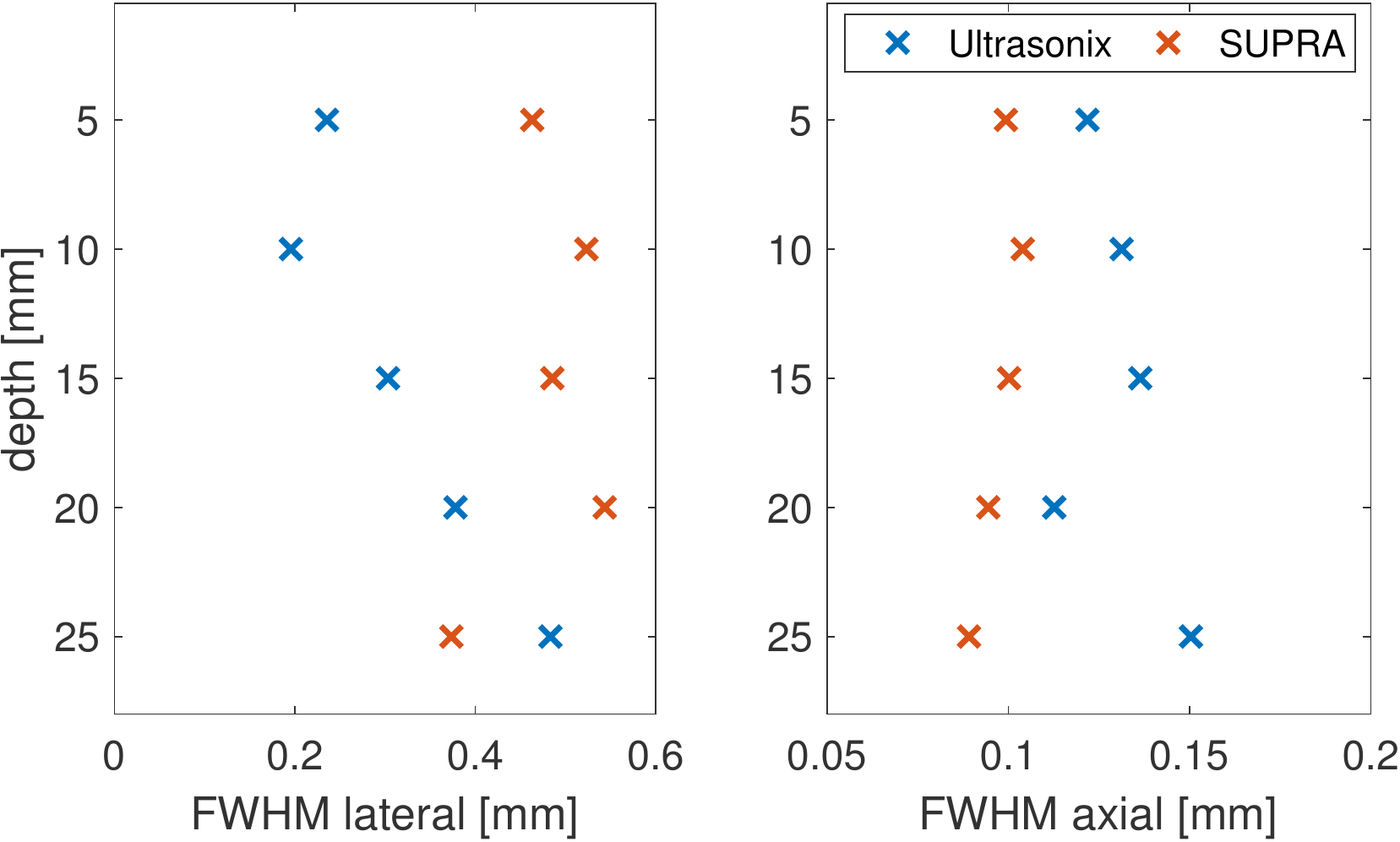}
	%	\label{fig:psf-comparison-b}
	%	}
	\caption{\shortcaption{Point spread functions} (PSFs) measured for the Ultrasonix and SUPRA beamformers with a linear probe. The top row shows exemplary PSFs at \SI{20}{mm} depth and \SI{50}{dB} dynamic range, the bottom row measurements of the full-width-half-maximum (FWHM) in lateral and axial directions.
	\label{fig:psf-comparison}
	}
	%\end{tabular}
\end{figure}

Given the Dirac-like reflector, images representing the PSFs at depths 5, 10, 15, 20 and \SI{25}{mm} are retrieved.
\autoref{fig:psf-comparison} shows in the top row the PSFs of both systems at \SI{20}{mm} and \SI{50}{dB}.
While the lateral extent of the  Ultrasonix PSF is smaller than that of SUPRA combined with the Cephasonics system and probe, its height is considerably larger.
This becomes even more clear in the bottom row of \autoref{fig:psf-comparison}, where the lateral PSF full-width-half-maximum (FWHM) of SUPRA is larger in most depth those of the Ultrasonix system (bottom left), whereas the the axial FWHM of SUPRA is smaller for all depths.

When considering an ultrasound imaging pipeline as a linear system, the PSF characterizes the system-response to a Dirac-pulse. With that property, it also serves as one measure of imaging resolution.
Considering the findings w.r.t. the PSFs of both systems, this leads to the conclusion, that the lateral resolution of the Ultrasonix pipeline (up to the beamformer) is higher than that of SUPRA with Cephasonics, while the axial resolution of SUPRA is higher.

This result only partially agrees with the observations from \autoref{subsec:qualitative}, where a more pronounced blurring could be observed for SUPRA, even in axial direction. This difference can be explained by the fact that the qualitative evaluation takes the complete imaging pipeline into account, while the PSF evaluation only considers the pipeline up to the beamforming.
Overall, however, it shows that SUPRA can provide comparable image quality, following the purely software-based approach.

\subsection{Performance}
\label{subsec:eval-performance}
In the following, we present a run-time analysis of a SUPRA pipeline consisting of beamformer, envelope detection, log compression and scan conversion.
We performed this evaluation on a number of computers
\begin{itemize}
    \item Dedicated workstation with a NVIDIA GeForce GTX 1080 / \SI{8}{GB} (Ubuntu Linux 14.04, Intel Xeon E5 - 1660 v4, \SI{3.2}{GHz}, 8 core, \SI{32}{GB} RAM)
    \item Notebook with a NVIDIA GeForce GTX 960M / \SI{2}{GB} (Windows 10, Dell XPS 15 9550, Intel Core i7 6700 HQ, \SI{2.6}{GHz}, 4 core, \SI{16}{GB} RAM)
    \item Jetson TX2, an embedded SoC with a total power target of \SI{15}{W}, includes a NVIDIA GPU sharing the main memory (Ubuntu Linux 16.04, ARM A57, \SI{2.0}{GHz}, 4 core, \SI{8}{GB} RAM)
\end{itemize}

As stated before, one of the design goals of SUPRA is its interactive use, consequently limiting the run-time of all nodes of a pipeline.
\autoref{tab:run-time} shows the node and pipeline run-times we observed in milliseconds on the different hardware configurations. Note that SUPRA used previously recorded raw channel data as input for the beamforming on the Windows laptop and the ARM SoC.
From this table a number of observations are noteworthy.
\begin{table}[htb]
	\footnotesize{
	\centering
	\begin{tabular}{cr|rrrr|r}
        & Scanlines & Beamformer & Envelope & Log comp. & Scan conv. & Total \\
        \hline
        \multicolumn{7}{l}{Jetson TX2 SoC, Integrated GPU / \SI{8}{GB} shared}\\
        \hline
        \multirow{ 4}{*}{\rotatebox[origin=c]{90}{2D linear}} &
          (64 / 1) &   \meanStd{7.02}{0.89} &  \meanStd{9.00}{1.31} & \meanStd{0.41}{0.16} & \meanStd{20.91}{2.88} & \meanStd{37.64}{8.22} \\
        & (64 / 2) &  \meanStd{11.21}{1.09} & \meanStd{12.80}{1.47} & \meanStd{1.01}{0.24} & \meanStd{21.45}{2.19} & \meanStd{46.66}{5.31} \\
        & (128 / 1) & \meanStd{11.09}{0.48} & \meanStd{11.32}{0.96} & \meanStd{0.75}{0.14} & \meanStd{19.68}{1.19} & \meanStd{43.43}{5.30} \\
        & (128 / 2) & \meanStd{20.27}{0.57} & \meanStd{16.80}{1.10} & \meanStd{0.66}{0.21} & \meanStd{19.27}{0.80} & \meanStd{57.44}{6.40} \\
        \hline
        \multicolumn{7}{l}{Notebook, NVIDIA GeForce GTX 960M / \SI{2}{GB}}\\
        \hline
        \multirow{ 4}{*}{\rotatebox[origin=c]{90}{2D linear}} &
          (64 / 1)  & \meanStd{3.35}{0.22} &  \meanStd{5.08}{0.64} & \meanStd{0.13}{0.03} & \meanStd{6.68}{0.26} & \meanStd{15.25}{1.01} \\
        & (64 / 2)  & \meanStd{4.97}{0.28} &  \meanStd{6.60}{1.00} & \meanStd{0.70}{0.22} & \meanStd{7.03}{0.45} & \meanStd{19.28}{2.07} \\
        & (128 / 1) & \meanStd{5.39}{0.23} &  \meanStd{9.28}{1.31} & \meanStd{0.80}{0.17} & \meanStd{7.18}{0.32} & \meanStd{22.70}{2.01} \\
        & (128 / 2) & \meanStd{9.04}{0.24} & \meanStd{11.04}{0.90} & \meanStd{0.74}{0.19} & \meanStd{7.14}{0.26} & \meanStd{28.15}{2.65} \\
        \hline
        \multicolumn{7}{l}{Dedicated workstation, NVIDIA GeForce GTX 1080 / \SI{8}{GB}}\\
        \hline
        \multirow{ 4}{*}{\rotatebox[origin=c]{90}{2D linear}} &
          (64 / 1)  & \meanStd{1.54}{0.21} & \meanStd{1.68}{0.19} & \meanStd{0.08}{0.02} & \meanStd{2.06}{0.19} & \meanStd{5.37}{0.44} \\
        & (64 / 2)  & \meanStd{0.93}{0.04} & \meanStd{0.98}{0.17} & \meanStd{1.10}{0.19} & \meanStd{1.23}{0.03} & \meanStd{4.24}{0.47} \\
        & (128 / 1) & \meanStd{1.03}{0.03} & \meanStd{0.98}{0.17} & \meanStd{1.15}{0.19} & \meanStd{1.22}{0.03} & \meanStd{4.38}{0.49} \\
        & (128 / 2) & \meanStd{1.68}{0.03} & \meanStd{1.98}{0.29} & \meanStd{0.08}{0.03} & \meanStd{1.23}{0.02} & \meanStd{5.00}{0.48} \\[0.5em]
        \hline
        \multicolumn{2}{l|}{3D phased} & \meanStd{11.15}{0.65} & \meanStd{2.49}{0.28} & \meanStd{0.08}{0.03} & \meanStd{13.79}{0.31} & \meanStd{27.49}{1.04}
    \end{tabular}
    }
    \caption{\shortcaption{Observed pipeline and node run-times $[\text{ms}]$} for 2D and 3D imaging (mean and standard deviation). The 2D imaging was performed in different scanline configurations ranging from 64 reconstructed scanlines without multiline receive (64 / 1) to 255 scanlines reconstructed from data of 128 transmit events (128 / 2) scanlines with depth \SI{45}{mm} and image resolution \SI{0.0225}{mm} isotropic. The 3D pipeline was parametrized as described in \autoref{subsec:qualitative} ($32 \times 16$ scanlines, field of view \SI{60}{\degree}, depth \SI{70}{mm}, image resolution \SI{0.175}{mm} isotropic) 
	\label{tab:run-time}}
\end{table}
The benchmarks on the NVIDIA Jetson TX2 show that a pure software 2D ultrasound pipeline can be executed on an embedded device with reasonable frame rates. This has the potential to enable mobile US-systems based on software beamforming.
It can be seen, that the log compression and scan conversion nodes exhibit only limited variation for the different scanline configurations. This is caused by two circumstances: While the log compression run-time is governed by the CUDA management overhead as it performs only limited computations, the run-time of the scan-conversion depends mostly on the resolution and size of the output.

As the GPU in the tested laptop is significantly faster than that present in the Jetson TX2, it is not surprising that the node run-times are lower.
On the dedicated workstation with a NVIDIA GTX 1080 it is clear that even the 2D beamforming is limited by CUDA management operations as its run-time is not influnced by the scanline configuration. The overall 2D pipeline should thus be able to perform significantly faster than \SI{100}{Hz}.

In addition to the 2D pipeline profiled on all three machines, we executed a 3D pipeline on the dedicated workstation and measured its run-time as well.
Although the number of scanlines was only twice as large as with the largest 2D pipeline, the beamforming took significantly more time. This is caused by the increased number of raw channels (384 vs. 64) and the resulting need to take those into account during beamforming. It can furthermore be observed, that the 3D scan conversion requires more time, which is caused by the addition of a whole dimension to its output.
Due to the significant memory requirements and number of neccessary operations of the 3D pipeline, we did not execute it on the laptop and the Jetson TX2.

This run-time analysis is shows that a purely software based US pipeline as implemented with SUPRA can in fact be used for real-time imaging, even on commodity graphics hardware.
\section{Conclusion}
\label{sec:conclusion}

In this work, we introduced an open source software based ultrasound processing framework that has the potential to make fundamental ultrasound imaging research accessible. The replacement of current hardware implementations of processing pipelines with software facilitates improvements and customizations which, so far, required specialized personnel and extensive development periods.
The platform allows for agile extensions and at the same time enables research by improving reproducibility.
The base framework, supporting a standard processing pipeline, permits flexible developments through its modular design, such that specialized solutions can be used in place of baseline algorithms.
With access to all intermediate data streams and the possibility of modifications on all stages, existing methods could be modified to include specific data that has not been considered until now.

With the performed evaluation, we demonstrate that the processing steps implemented provide image quality that is comparable to a clinical system.
Additionally, the run-time analysis proves the real-time capabilities of SUPRA.
It is also notable that the license under which SUPRA is distributed allows for both research and commercial oriented development. 
To this end, we aim at creating a community around this platform to support its future development and extension.
Thus, we encourage research groups to evaluate the SUPRA framework and contribute to its growth.

%\begin{acknowledgements}
%If you'd like to thank anyone, place your comments here
%and remove the percent signs.
%\end{acknowledgements}

% BibTeX users please use one of
%\bibliographystyle{spbasic}      % basic style, author-year citations
%\bibliographystyle{spmpsci}      % mathematics and physical sciences
%\bibliographystyle{spphys}       % APS-like style for physics

\bibliographystyle{amsplain} 
\bibliography{bib}   % name your BibTeX data base

\providecommand{\bysame}{\leavevmode\hbox to3em{\hrulefill}\thinspace}
\providecommand{\MR}{\relax\ifhmode\unskip\space\fi MR }
% \MRhref is called by the amsart/book/proc definition of \MR.
\providecommand{\MRhref}[2]{%
  \href{http://www.ams.org/mathscinet-getitem?mr=#1}{#2}
}
\providecommand{\href}[2]{#2}
\begin{thebibliography}{10}

\bibitem{averkiou2000}
M.~A. Averkiou, \emph{Tissue harmonic imaging}, 2000 IEEE Ultrasonics
  Symposium. Proceedings. An International Symposium, vol.~2, Oct 2000,
  pp.~1563--1572.

\bibitem{bluemel2016fusion}
Christina Bluemel, Gonca Safak, Andreas Cramer, Achim W{\"o}ckel, Anja
  Gesierich, Elena Hartmann, Jan-Stefan Schmid, Franz Kaiser, Andreas~K. Buck,
  and Ken Herrmann, \emph{Fusion of freehand {SPECT} and ultrasound: First
  experience in preoperative localization of sentinel lymph nodes}, European
  Journal of Nuclear Medicine and Molecular Imaging \textbf{43} (2016), no.~13,
  2304--2312.

\bibitem{Jeong2015}
Mok~Kun Jeong and Sung~Jae Kwon, \emph{{Estimation of side lobes in ultrasound
  imaging systems}}, Biomedical Engineering Letters \textbf{5} (2015), no.~3,
  229--239.

\bibitem{Krissian2007}
Karl Krissian, Carl~Fredrik Westin, Ron Kikinis, and Kirby~G. Vosburgh,
  \emph{{Oriented speckle reducing anisotropic diffusion}}, IEEE Transactions
  on Image Processing \textbf{16} (2007), no.~5, 1412--1424.

\bibitem{Lasso2014}
Andras Lasso, Tamas Heffter, Adam Rankin, Csaba Pinter, Tamas Ungi, and Gabor
  Fichtinger, \emph{{PLUS}: {O}pen-source toolkit for ultrasound-guided
  intervention systems}, IEEE Transactions on Biomedical Engineering
  \textbf{61} (2014), no.~10, 2527--2537.

\bibitem{riva20173d}
Marco Riva, Christoph Hennersperger, Fausto Milletari, Amin Katouzian, Federico
  Pessina, Benjamin Gutierrez-Becker, Antonella Castellano, Nassir Navab, and
  Lorenzo Bello, \emph{{3D} intra-operative ultrasound and {MR} image guidance:
  pursuing an ultrasound-based management of brainshift to enhance
  neuronavigation}, International Journal of Computer Assisted Radiology and
  Surgery \textbf{12} (2017), no.~10, 1711--1725.

\bibitem{Rodriguez-Molares2017}
Alfonso Rodriguez-Molares, Ole Marius~Hoel Rindal, Olivier Bernard, Arun Nair,
  Muyinatu A.~Lediju Bell, Herv{\'{e}} Liebgott, Andreas Austeng, and Lasse
  L{\o}vstakken, \emph{{The UltraSound ToolBox}}, 2017.

\bibitem{Shattuck1984}
D~P Shattuck, M~D Weinshenker, S~W Smith, and O~T von Ramm, \emph{{Explososcan:
  a parallel processing technique for high speed ultrasound imaging with linear
  phased arrays.}}, The Journal of the Acoustical Society of America
  \textbf{75} (1984), no.~4, 1273--1282.

\bibitem{tanter2014ultrafast}
M.~Tanter and M.~Fink, \emph{Ultrafast imaging in biomedical ultrasound}, IEEE
  Transactions on Ultrasonics, Ferroelectrics, and Frequency Control
  \textbf{61} (2014), no.~1, 102--119.

\bibitem{Zettinig2015}
Oliver Zettinig, Amit Shah, Christoph Hennersperger, Matthias Eiber, Christine
  Kroll, Hubert K{\"u}bler, Tobias Maurer, Fausto Milletar{\`i}, Julia
  Rackerseder, Christian Schulte~zu Berge, Enno Storz, Benjamin Frisch, and
  Nassir Navab, \emph{Multimodal image-guided prostate fusion biopsy based on
  automatic deformable registration}, International Journal of Computer
  Assisted Radiology and Surgery \textbf{10} (2015), no.~12, 1997--2007.

\end{thebibliography}

% Non-BibTeX users please use
% \begin{thebibliography}{}
%
% and use \bibitem to create references. Consult the Instructions
% for authors for reference list style.
%
% \bibitem{RefJ}
% Format for Journal Reference
% Author, Article title, Journal, Volume, page numbers (year)
% Format for books
% \bibitem{RefB}
% Author, Book title, page numbers. Publisher, place (year)
% etc
% \end{thebibliography}

\end{document}